\documentclass[10pt,twocolumn,letterpaper]{article}

\usepackage{iccv}
\usepackage{times}
\usepackage{epsfig}
\usepackage{graphicx}
\usepackage{amsmath}
\usepackage{subcaption,booktabs}

\usepackage{url}
\usepackage{array}
\usepackage{multirow}
\usepackage{listings}
\usepackage{sidecap}
\DeclareMathOperator*{\argmax}{arg\,max}

\usepackage{amssymb}
\newcommand{\mypara}[1]{\noindent{\bf{#1}}}
\usepackage{xcolor}

\newcommand{\blue}[1]{\textcolor{blue}{#1}}

\newcommand{\ourmodel}{\texttt{Waffle}CLIP\xspace}

\usepackage[colorlinks=True]{hyperref}

\iccvfinalcopy 

\def\iccvPaperID{2713} 

\ificcvfinal\fi

\begin{document}

\title{Waffling around for Performance: Visual Classification with \\ Random Words and Broad Concepts}

\author{Karsten Roth$^{1,*}$, Jae Myung Kim$^{1,*}$, A. Sophia Koepke$^{1}$, Oriol Vinyals$^{2}$, Cordelia Schmid$^{3}$, Zeynep Akata$^{1,4}$\\
\small{$^{1}$University of Tübingen, Tübingen AI Center, $^{2}$Google DeepMind,}\\ 
\small{$^{3}$Inria, Ecole normale sup\'erieure, CNRS, PSL Research University, $^{4}$MPI for Intelligent Systems}\\
\small{$^*$equal contribution}}

\maketitle
\ificcvfinal\thispagestyle{empty}\fi

\begin{abstract}
The visual classification performance of vision-language models such as CLIP has been shown to benefit from additional semantic knowledge from large language models (LLMs) such as GPT-3. 
In particular, averaging over LLM-generated class descriptors, e.g.\ ``waffle, \textit{which has a round shape}'', can notably improve generalization performance.
In this work, we critically study this behavior and propose \texttt{Waffle}CLIP, a framework for zero-shot visual classification which simply replaces LLM-generated descriptors with random character and word descriptors. \textbf{Without} querying external models, we achieve comparable performance gains on a large number of visual classification tasks. This allows \texttt{Waffle}CLIP to both serve as a low-cost alternative, as well as a sanity check for any future LLM-based vision-language model extensions.
We conduct an extensive experimental study on the impact and shortcomings of additional semantics introduced with LLM-generated descriptors, and showcase how - if available - semantic context is better leveraged by querying LLMs for high-level concepts, which we show can be done to jointly resolve potential class name ambiguities. 
Code is available here: \href{https://github.com/ExplainableML/WaffleCLIP}{https://github.com/ExplainableML/WaffleCLIP}.  
\end{abstract}

\section{Introduction}\label{sec:introduction}

\begin{figure}[t]
    \centering
    \includegraphics[width=\linewidth]{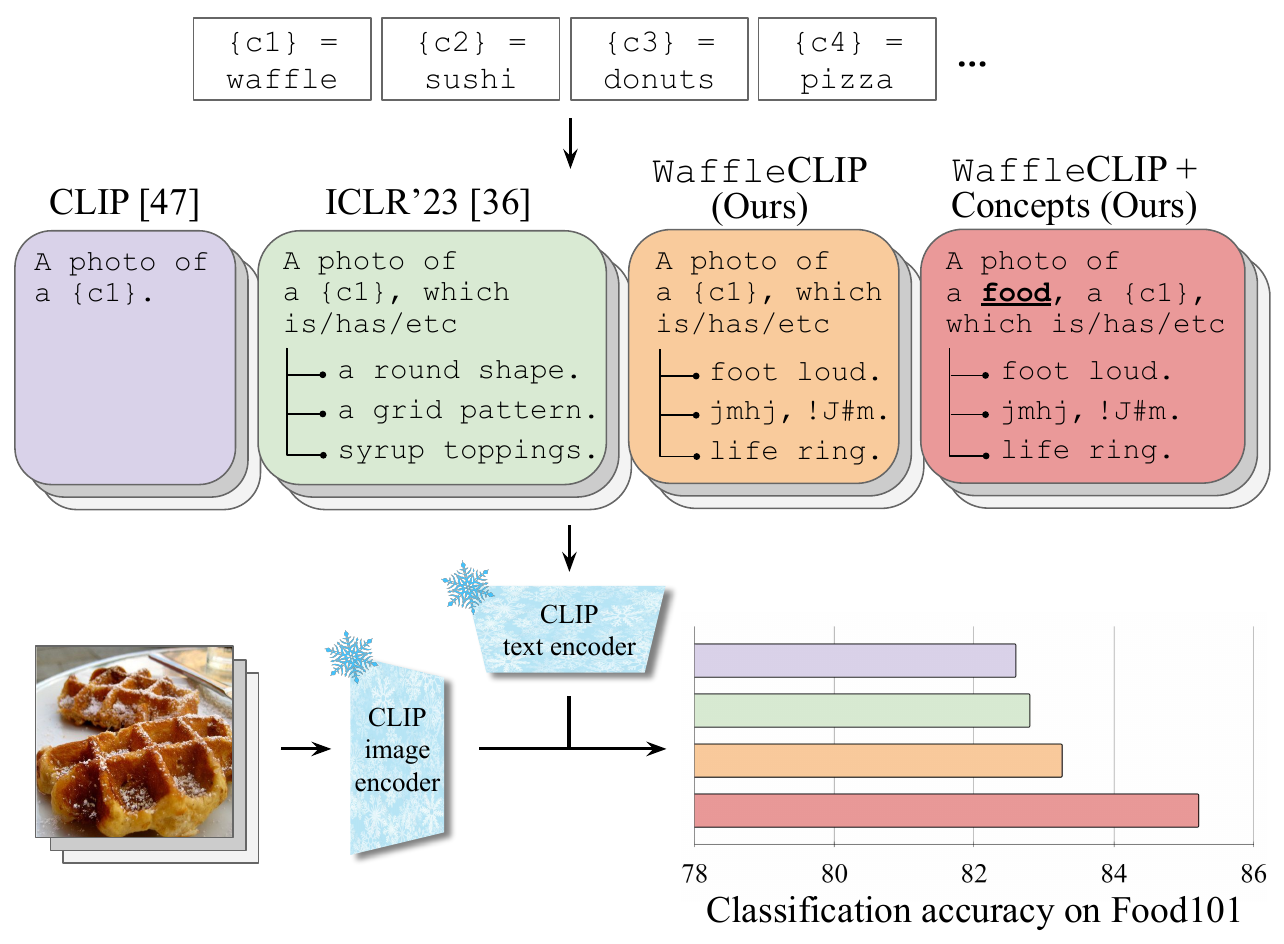} 
    \vspace{-10pt}
    \caption{Substituting GPT-3 generated fine-grained descriptors with random word or character sequences yields competitive performance. High-level concepts further resolve classname ambiguities for additional gains.}
    \label{fig:teaser}
    \vspace{-10pt}
\end{figure}

Task-specific tuning of natural language prompts \cite{prompt2,coop,prompt3,prompt4} can improve the performance of large vision-language models (VLMs)~\cite{clip}. However, if the model does not have access to additional training data, i.e.\ in the zero-shot setting, this is not an option.
Instead, a promising alternative \cite{chils,cupl,dclip} is querying large language models (LLMs) to provide additional semantic context to enrich class representations. Extending classnames with fine-grained class descriptors generated by GPT-3~\cite{gpt3} and minimal human intervention has experimentally shown to boost results \cite{dclip,cupl}, for instance with class-based descriptors on top of classnames, e.g.\ \textit{a round shape} for \textit{waffle} \cite{dclip}.

However, close inspection of GPT-3 generated semantic cues indicates a high degree of diversity, limited visual relevance, and ambiguity \cite{dclip}. For instance, multiple descriptors can be assigned to the same class despite likely not co-occurring, e.g.\ \textit{``steamed''} and \textit{``fried''}, 
or non-visual attributes might be mentioned, e.g.\ \textit{``a sour and spicy smell''}, or the class interpretation might be ambiguous, e.g.\ \textit{``webbed feet''} for \textit{``Peking duck''} as a food item. 
Hence, the underlying drivers of performance improvements when using generated fine-grained class descriptors are unclear. 

To understand these performance gains, we first show that each set of class-specific GPT-3 generated descriptors can be replaced with a fixed set of randomly selected, class-independent descriptors while still retaining similar benefits in performance.
Motivated by this observation, we take this one step further and propose \ourmodel, named after \textit{waffling around} the class name, that replaces the LLM-generated fine-grained descriptors, e.g.\ \textit{a round shape, a grid pattern}, with random words (e.g.\ \textit{"foot loud"}) or character lists (e.g.\ \textit{"jmhj, !J\#m"}) based on average class name length and word counts (cf.~Figure~\ref{fig:teaser}).
As \ourmodel does not require access to LLMs for additional context (unlike e.g.~\cite{dclip,chils,cupl,susx}), it remains \textit{inherently zero-shot}. Consequently, it also serves as an important sanity check for future methods utilizing external model queries.

Naturally, the convincing performance of \ourmodel across benchmarks raises questions regarding the true benefits of additional semantics introduced by LLM-generated descriptors. 
We provide answers with extensive experiments, showcasing that semantic descriptors produced by LLMs offer a \textit{structurally} different and \textit{complementary} impact on the classification behavior. However, we find this not to be fully driven by additionally introduced semantics, but rather a different form of structured noise ensembling. 

Instead, we show that actual semantic context is better introduced through coarse-grained, high-level concepts.
Given access to external LLMs, we suggest a query mechanism for GPT-3 to automatically generate these concepts (e.g.\ \textit{food} for \textit{waffle, peking duck}), while jointly resolving issues of context-dependent class label ambiguity.

In summary, our contributions are:
\textbf{1)} We motivate and propose \ourmodel to use random character and word descriptors to enhance the semantic retrieval process in VLMs (particularly CLIP);
\textbf{2)} we demonstrate that \ourmodel yields comparable zero-shot classification performances at lower cost compared to methods reliant on LLM-generated descriptors, thus also serving as an important sanity check for future models;
\textbf{3)} we extensively study the semantic context introduced through LLM-generated descriptors and propose (automatically extracted) high-level LLM-generated concepts as an alternative for better use of semantics while tackling classname ambiguities.

\section{Related Work}
Image classification with VLMs such as CLIP~\cite{clip} has gained popularity particularly in low-data regimes.
As input prompts have a significant impact on the performance, recent research has focused on the exploration of learnable prompts for the text encoder~\cite{coop,cocoop,distcoop,ttpt}, the visual encoder~\cite{bahng2022exploring,chen2022understanding,wu2022unleashing,loedeman2022prompt} or for both encoders jointly~\cite{xing2022class}. 

Alternatively, synthetic images generated using available classnames can support corresponding image classification~\cite{susx,bansal2023leaving,he2023is}. 
In contrast, we do not tune prompts or query external image generation methods, but propose to use prompts containing random characters or words to enhance the zero-shot capabilities of VLMs. 

\mypara{Adding external knowledge to language prompts.} 
Recently, multiple works have shown how LLMs can be leveraged to obtain more effective prompts. 
\cite{cupl,naeem2022i2mvformer,mao2023doubly} utilized GPT-3~\cite{gpt3} to produce and study lengthy, descriptive sentences that articulate the visual concepts for each category, while
\cite{chils} generated semantic hierarchies to identify subclasses of categories for zero-shot class prediction. \cite{dclip} used multiple fine-grained LLM-generated class descriptors, which enhance accuracy and appear to provide interpretability by assigning weights to each descriptor. 

Similarly, different kinds of descriptions have been used for image classification tasks, by manually crafting descriptions~\cite{reed2016learning,he2017fine}, or by utilizing external databases based on e.g. Wikipedia~\cite{elhoseiny2017link,paz2020zest,naeem2022i2dformer,choudhury2021curious},
the WordNet hierarchy~\cite{miller1998wordnet,klite,langdml}, or the ImageNet-Wiki~\cite{bujwid2021large}. 

Whilst external knowledge from LLMs can be valuable, in this work we show that one can match the image classification performance gains of using fine-grained LLM-generated descriptors with randomly sampled characters and words as class descriptors. In addition, we find that if semantic context is available through LLMs, it is better integrated through high-level context (c.f.\ also \cite{dunlap2023using}), for which we provide an automatic extraction mechanism.

\mypara{Noise augmentation.}
Data augmentation through noise is known to enhance the performance and robustness of model training for a variety of tasks and domains~\cite{shorten2019survey,feng2021survey}. In the language domain, noise can be incorporated in the embedding or input space.
For instance, \cite{sun2020mixup,chen2020mixtext,hao2023mixgen} used linguistic embedding space augmentations inspired by mixup~\cite{mixup}, and \cite{cheng2018towards} added Gaussian embedding space noise.
Augmentation through input space noise has been performed at the word-~\cite{kobayashi2018contextual,wei2019eda}, token-~\cite{wei2019eda} or character-level~\cite{csahin2022augment,heigold2018robust,belinkov2018synthetic,niu2020evaluating}.

For character-level noise augmentation, characters are randomly substituted, added, or removed \cite{heigold2018robust,belinkov2018synthetic,niu2020evaluating}.
In all cases, these augmentation are used to prevent overfitting \textit{during training}. Instead, our approach utilizes character- and word-level language augmentation to perturb the class prompts for improved zero-shot image classification.

\begin{table*}[t]
\centering\resizebox{\linewidth}{!}{
\begin{tabular}{l|llllllll|l}
\toprule
\textbf{ViT-B/32} & ImageNetV2 & ImageNet & CUB200 & EuroSAT & Places365 & Food101 & Oxford Pets & DTD & \textbf{Avg}\\
\hline
CLIP~\cite{clip} &  54.71 & 62.01 & 51.28 & 40.78 & 39.12 & 82.59 & 85.06 & 43.18 & 57.34 \\
DCLIP~\cite{dclip} &  \textbf{55.82} & \textbf{63.12} & 52.47 & \textbf{43.29} & 40.47 & 82.79 & 86.54 & \textbf{43.99} & \blue{\textbf{58.56}}\\
\hline
DCLIP (same, 1x)  & 55.47 \small$\pm0.24$ & 62.89 \small$\pm0.19$ & 52.64 \small$\pm0.28$ & 39.74 \small$\pm2.69$ & 40.29 \small$\pm0.47$ & 83.82 \small$\pm0.48$ & 87.04 \small$\pm0.27$ & 43.35 \small$\pm0.41$ & 58.16 \small$\pm1.01$\\
DCLIP (same, 2x) & 55.75 \small$\pm0.21$ & 63.10 \small$\pm0.19$ & \textbf{52.72} \small$\pm0.23$ & 39.73 \small$\pm1.66$ & \textbf{40.61} \small$\pm0.22$ & \textbf{84.01} \small$\pm0.23$ & \textbf{87.10} \small$\pm0.14$ & 43.29 \small$\pm0.22$ & 58.29 \small$\pm0.62$\\
\bottomrule
\end{tabular}
}
\caption{\textbf{Motivating random class descriptors.} Comparing CLIP~\cite{clip} and the GPT-descriptor-extended CLIP~\cite{dclip} (DCLIP) with the same set of randomly sampled descriptors for each class, where the set size is either the average number of descriptors per class in DCLIP (\textit{same, 1x}), or twice that (\textit{same, 2x}). 
A random set of descriptors per class can match or even outperform DCLIP 
across backbone architectures (results for ViT-L/14 and ResNet50 are included in the suppl.\ material) confirming that randomized prompt averaging leads to higher performance.}
\label{tab:motivation}
\end{table*}
\section{Method}\label{sec:method}

We first describe image classification using class descriptors following \cite{dclip} (\S\ref{subsec:method_0}), before motivating and explaining our LLM-free, random semantic descriptor alternative \texttt{Waffle}CLIP (\S\ref{subsec:method_1}). Finally, if LLMs are available, we highlight a simple extension to incorporate semantics while jointly resolving ambiguities with automatically extracted high-level semantic concepts (\S\ref{subsec:method_2}).

\subsection{Image classification with class descriptors}\label{subsec:method_0}
Given target categories $C$ and a query image $x$, the zero-shot image classification protocol used in CLIP~\cite{clip} defines the classification problem as nearest neighbour retrieval:
\begin{equation}
    \tilde{c} = \argmax_{c\in C} s(\phi_I(x), \phi_L(f(c))),
\end{equation}
with prompt $f(c)=$ \texttt{"A photo of a \{c\}."} and image and language encoder $\phi_I$ and $\phi_L$. 
To improve the retrieval process, \cite{dclip} converts the simple class-embedding retrieval to a dictionary-based one, where a class $c$ is associated with a set of descriptors $D_c$: \texttt{"\{\textit{c}\} which (is/has/etc) \{descriptor\}."} with e.g.\ \texttt{\textit{c} = "waffle"} and \texttt{descriptor = "a round shape"}. Given $D_c$ for classes \textit{c}, classification is reformulated as
\begin{equation}\label{eq:cls}
    \argmax_{c\in C} \frac{1}{|D_c|}\sum_{d\in D_c}s(\phi_I(x), \phi_L(d)),
\end{equation}
which defines the similarity between the image $x$ and class $c$ as the average similarity to all its descriptor variants. We abbreviate this descriptor-based extension of CLIP as \textit{DCLIP}.

\subsection{\texttt{Waffle}CLIP}\label{subsec:method_1}
DCLIP \cite{dclip} \footnote{DCLIP \cite{dclip} reports improvements over CLIP by using the phrase \texttt{"\{\textit{c}\}, which (is/has/etc) \{descriptor\}."} instead of \texttt{"A photo of \{\textit{c}\}, which (is/has/etc) \{descriptor\}."} as suggested and studied in the original CLIP paper. For fair comparison with existing baseline CLIP results, we thus utilize the latter, but find similar behaviour for other prompt structures.} requires external LLMs for descriptors that convert the single-class matching problem to one over an ensemble of fine-grained class representations.  

\subsubsection{Motivation} 
We observe that LLM-generated class descriptors reveal high diversity, limited visual relevance, and ambiguity.
From a conceptual perspective, this makes it hard to pin down the precise benefits of generated class descriptors used e.g.\ in \cite{cupl} or \cite{dclip}.
\begin{figure*}[t]
    \centering
    \includegraphics[width=1\linewidth]{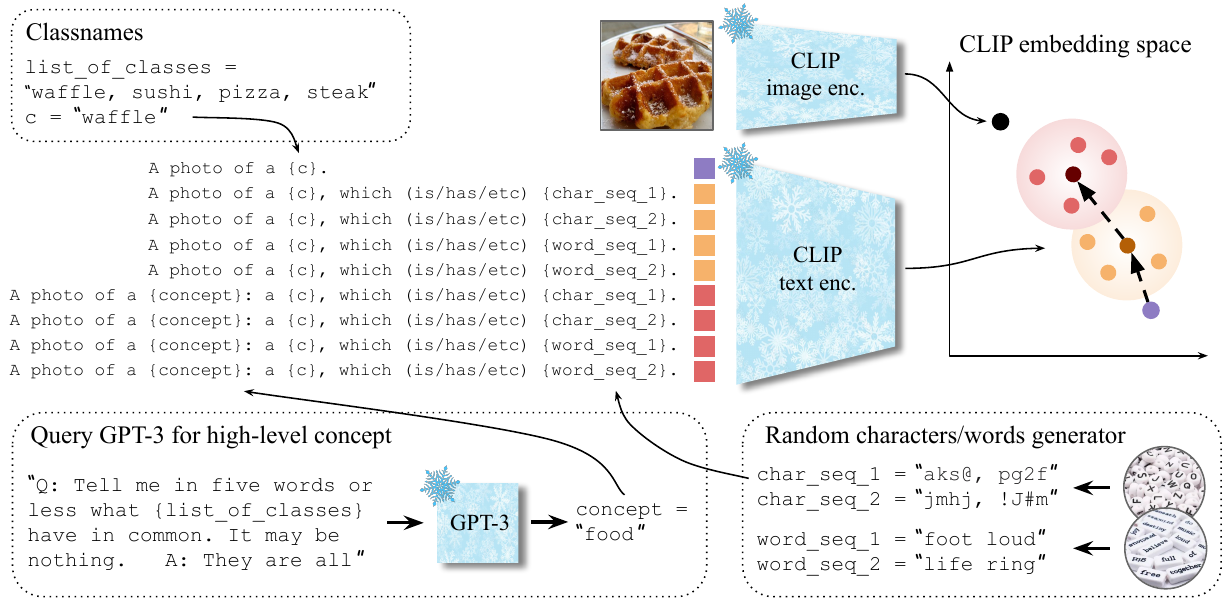} 
    \vspace{-10pt}
    \caption{\textbf{Visual classification with \ourmodel using random characters/words.} 
    \ourmodel utilizes a collection of prompt variations by simply injecting character-level and word-level noise around the classname (\textit{top left}, \textit{orange}). Simple averaging consequently raises the robustness of the extracted semantics and the corresponding retrieval process (\textit{top right}), leading to notable gains in the open-vocabulary classification performance using Vision-Language models such as CLIP, while approximating the performance improvements gained by e.g. querying external Large Language models for additional semantic descriptors. In addition, we show that if given access to external LLMs, \ourmodel can be further enhanced by adding a high-level concept descriptor in the prompt (\textit{red}).}
    \label{fig:main}
\end{figure*}
To understand a possible driver of performance improvements, we conduct a simple experimental study, shown in Tab.~\ref{tab:motivation}. We take all available LLM-generated descriptors for a dataset from \cite{dclip}, sample a small set of descriptors where the cardinality of the set is 
the average number of descriptors per class used in DCLIP, and assign this same set of random descriptors to every class, i.e.\ \textit{DCLIP (same, 1x)}. 

This shows a close match to DCLIP (e.g.\ $58.56\%$ and $58.16\%$ for ViT-B/32
in total average) and in parts even better performance (e.g.\ $0.83\%$
improvement in Food101 for ViT-B/32
). This reveals averaging over descriptor variations as one of the key drivers for performance.
The results further improve when increasing the number of random LLM-generated descriptors for each class (\textit{DCLIP (same, 2x)}, e.g.\ $58.16\%\rightarrow58.29\%$ for ViT-B/32
). 

Correspondingly, these results indicate that the role of additional descriptor semantics is likely overestimated, especially when uncurated descriptors are used.
Building on the benefits of averaging over various prompt variants to extract a better semantic representation estimate of an associated class, we investigate whether fully randomized prompt descriptors can provide similar benefits, \textbf{without} querying external LLMs.\\ 

\subsubsection{\texttt{Waffle}CLIP Method} This motivates \texttt{Waffle}CLIP, an \textit{LLM-free} descriptor alternative that uses simple randomized descriptors. 
In particular, we populate $D_c$ with class-independent, random word sequences or random character lists, with a fixed number of characters per word $l_w$, and a fixed number of words $n_w$. For example, \ $l_w = 4$ and $n_w = 2$ for \texttt{char\_seq\_1 = "aks@, pg2f"} in Fig.~\ref{fig:main}.
To avoid introducing hyperparameters, we leverage a simple heuristic where the average number of words and average number of characters per word in the provided class labels determines $l_w$ and $n_w$.
As a result, this converts the standard CLIP input prompt \texttt{"A photo of a \{c\}."} into \texttt{"A photo of a \{c\}, which (is/has/etc) \{random\_sequence\}."}, where we follow the extension structure used in \cite{dclip}.

\subsection{Better semantics and reduced ambiguity via high-level concepts}\label{subsec:method_2}
Due to the limited impact of additional semantics introduced by fine-grained descriptors (c.f.\ \S\ref{subsec:method_1}), we propose an alternative way of querying LLMs that does not require averaging across multiple descriptors and simultaneously addresses the issue of class ambiguities.
Therefore, we suggest taking a step back and searching not for additional class details, but instead for higher-level commonalities \textit{between} the classes, akin to the use of class hierarchies in image classification \cite{guo2018cnn}. Understanding commonalities between multiple target classes can help resolve ambiguities. If the class \texttt{"boxer"} is seen in the context of animal classification, it likely refers to the animal instead of a human athlete. 

We propose to automatically produce such high-level concepts by using available class names (or subsets if the class count exceeds the maximum LLM input sequence length) $C_\mathcal{D}$ for a dataset $\mathcal{D}$ and querying GPT-3~\cite{gpt3} with: 
\begin{center}
    \texttt{"Q: Tell me in five words or less what \{list\_of\_classes\} have in common. It may be nothing. A: They are all "}.
\end{center}
or small variants thereof. After extracting the shared \texttt{concept} through the corresponding LLM, simple filtering of concepts is executed to check if generated concepts are non-specific, namely \texttt{"Object"}, \texttt{"Thing"}, \texttt{"Verb"}, \texttt{"Adjective"}, \texttt{"Noun"}, or \texttt{"Word"}. If so, high-level concept guidance is omitted (this is only the case for three out of eleven visual classification benchmarks, see also \S\ref{subsec:experiments}). We then augment the default CLIP prompt to \texttt{"A photo of a \{concept\}: a \{c\}."} and for \ourmodel, the prompt is extended to \texttt{"A photo of a \{concept\}: a \{c\}, which (is/has/etc) \{random\_sequence\}."}
While the prompt style can likely be improved, this naive extension already offers remarkable benefits.


\section{Experiments}\label{subsec:experiments}
\begin{table*}[t]
\centering\resizebox{\linewidth}{!}{
\begin{tabular}{l|llllllll|l}
\toprule
\textbf{ViT-B/32} & ImageNetV2 & ImageNet & CUB200 & EuroSAT & Places365 & Food101 & Oxford Pets & DTD & \textbf{Avg}\\
\hline
CLIP~\cite{clip} & 54.71 & 62.01 & 51.28 & 40.78 & 39.12 & 82.59 & 85.06 & 43.18 & 57.34 \\
+ Concepts & $\downarrow$ & $\downarrow$ & 52.23 & 48.86 & 39.31 & 84.66 & 86.73 & $\downarrow$ & 58.96\\
\hline
DCLIP~\cite{dclip} &  55.82 & 63.12 & 52.47 & 43.29 & 40.47 & 82.79 & 86.54 & \textbf{43.99} & 58.56\\
\ourmodel (ours) & \textbf{55.92} \small$\pm0.08$ & \textbf{63.31} \small$\pm0.09$ & \textbf{52.38} \small$\pm0.12$ & 44.31 \small$\pm1.07$ & 40.56 \small$\pm0.07$ & 83.25 \small$\pm0.21$ & 85.70 \small$\pm0.25$ & 43.16 \small$\pm0.25$ & 58.57 \small$\pm0.41$\\
+ Concepts & $\downarrow$ & $\downarrow$ & 52.83 \small$\pm0.19$ & 48.51 \small$\pm0.70$ & 40.97 \small$\pm0.08$ & \textbf{85.21} \small$\pm0.06$ & 87.52 \small$\pm0.10$ & $\downarrow$ & 59.47 \small$\pm0.42$\\
+ GPT descr. + Concepts & $\downarrow$ & $\downarrow$ & \textbf{52.77} \small$\pm0.26$ & \textbf{51.64} \small$\pm0.25$ & \textbf{41.35} \small$\pm0.09$ & 84.87 \small$\pm0.05$ & \textbf{87.71} \small$\pm0.18$ & $\downarrow$ & \blue{\textbf{60.21}} \small$\pm0.20$\\
\bottomrule
\end{tabular}
}
\caption{\textbf{Image classification with \ourmodel} which extends input prompts with random word and character sequences and matches the performance of DCLIP \cite{dclip} using GPT-generated class descriptors. Additional semantic context through high-level concepts (\textit{+ Concepts}) can offer further boosts, particularly on benchmarks where classnames can be generic or ambiguous.  We further find that \ourmodel complements the use of GPT-generated descriptors (\textit{+ GPT descr.}). $(\downarrow)$ denotes same results as previous lines where high-level concept guidance is not applicable. For ViT-L/14 and RN50, see Supp.}
\label{tab:main}
\end{table*}
We first provide implementation details before comparing \texttt{Waffle}CLIP to DCLIP in \S\ref{subsec:random_descriptors}. 
Extending our observations in Tab.~\ref{tab:motivation}, we study the source of performance gains via LLM-generated descriptors (\S\ref{subsec:llm}) and present a better way for introducing semantics into the retrieval process while tackling semantic ambiguities with automatically extracted high-level concepts (\S\ref{subsec:highlevel}). 

In addition to that, \S\ref{subsec:add_benchmarks} provides further insights on other benchmarks not studied in e.g. \cite{dclip}, alongside benchmarks measuring out-of-distribution (OOD) generalization. \S\ref{subsec:promptensembles} showcases a comparison to hand-crafted prompt ensembles, and \S\ref{subsec:latent_noise} a study into the relationship of \ourmodel and latent space noise. \S\ref{subsec:ablations} then provides ablational studies to \ourmodel. Finally, further experiments and particularly experimental details and results are included in the supplementary materials.

\subsection{Experimental details} 
We utilize CLIP \cite{clip} as the underlying VLM for \ourmodel. As there is no direct cost associated with generating random character or word sequences, their number is only bounded by inference speed requirements (which is minimal as all respective language embeddings can be computed \textit{a priori} \cite{dclip}). However, we find diminishing returns for very high numbers (see also \S\ref{subsec:ablations}), and use 30 random descriptors per class (or 15 random character and word descriptor pairs) if not mentioned otherwise, with similar performance for both half or double the descriptor count (c.f.\ \S\ref{subsec:ablations}).
All experiments use PyTorch~\cite{pytorch} and are conducted on a single NVIDIA 3090Ti GPU. Wherever necessary, fine-grained LLM-generated descriptors are either taken from or generated following the codebase provided by \cite{dclip}. If not mentioned explicitly, all results involving \ourmodel are computed over seven random seeds.

\textbf{Benchmarks.} The datasets considered are (mostly from \cite{dclip}) ImageNet~\cite{imagenet} and ImageNetV2~\cite{imagenetv2}, CUB200-2011~\cite{cub200-2011} (fine-grained bird classification), EuroSAT~\cite{eurosat} (satellite image recognition), Places365~\cite{places365}, Food101~\cite{food101}, Oxford IIIT Pets~\cite{pets}, DTD (Textures, \cite{dtd}), Flowers102 \cite{flowers102}, FGVCAircraft \cite{fgcvaircraft}, and Stanford Cars \cite{cars196}.

\textbf{High-level concepts.} Following \S\ref{subsec:method_2}, the GPT-3 generated high-level concept for CUB200-2011 is \texttt{"Bird"}, \texttt{"Land Use"} for EuroSAT, \texttt{"Place"} for Places365, \texttt{"Food"} for Food101 and \texttt{"Breed"} for Oxford Pets. For additional benchmarks, extracted concepts are noted in section \S\ref{subsec:add_benchmarks}. For ImageNet (V2) and DTD, the concepts are too generic and thus filtered out (\texttt{"Object"}, \texttt{"Noun"}, or \texttt{"Adjective"}), with high-level guidance omitted.

\subsection{\ourmodel vs LLM-generated descriptors}\label{subsec:random_descriptors}

We start by analyzing the impact of randomization beyond fixed, randomized sets of fine-grained LLM-generated descriptors as done in Tab.~\ref{tab:motivation}, by instead using randomized character or word descriptors through our proposed \ourmodel. 
For that, we investigate visual classification accuracies across the eight diverse benchmarks studied in \cite{dclip} in Tab.~\ref{tab:main}, where we compare \ourmodel, which does not use any external LLMs, with DCLIP. 

We find that averaging over randomized descriptors yields performances comparable to or better than those obtained with LLM-generated fine-grained descriptors over a majority of studied datasets, with average performance being comparable: $58.56\%$ using DCLIP versus $58.57\%$ for \ourmodel with a ViT-B/32 backbone and (see supp. material) $69.14\%\rightarrow68.95\%$ for ViT-L/14, and $54.77\%\rightarrow54.20\%$ for ResNet50.
Beyond the inherently zero-shot nature of \ourmodel and ease of use, these results highlight that improved visual classification with pretrained VLMs does not require external LLMs, and further cements prompt averaging as a potential key driver behind DCLIP.

\subsection{Are descriptors from LLMs obsolete?}\label{subsec:llm}
Our results above question the benefits of LLM-generated fine-grained semantics, as averaging over fully randomized character and word sequences achieves comparable performance. But does that mean that there is no benefit in leveraging descriptors produced by LLMs?

\subsubsection{Impact of Descriptor Averaging} 
To better understand this, we extend our motivational experiments from Tab.~\ref{tab:motivation}. First, we look at what happens when not performing averaging over all class descriptor distances as in DCLIP, but instead choosing the maximum. If additional fine-grained semantics were indeed beneficial, selecting the most suitable one should similarly raise the performance. However, as Tab.~\ref{tab:max_dclip_main} reveals, performance actually drops, showing that the VLM cannot leverage the additional semantics to improve visual classification performance\footnote{This is potentially influenced by bag-of-words behavior of CLIP-like Vision-Language models as studied e.g. in \cite{yuksekgonul2023bow}. We leave the detailed analysis of this to future research.}. Instead, this again points to descriptor ensembling \textit{as the main driver in performance}.
\begin{table}[t]
\centering\resizebox{\linewidth}{!}{
\begin{tabular}{l|lllll}
\toprule
\textbf{ViT-B/32} & ImageNetV2 & ImageNet & CUB200 & EuroSAT & Places365\\
\hline
CLIP~\cite{clip} & 54.71 & 62.01 & 51.28 & 40.78 & 39.12 \\
\hline
DCLIP~\cite{dclip} (\textit{mean})&  \textbf{55.82} & \textbf{63.12} & \textbf{52.47} & \textbf{43.29} & \textbf{40.47}\\
DCLIP~\cite{dclip} (\textit{max})&  54.41   & 61.67   & 52.40   & 37.11   & 37.21\\
\bottomrule
\end{tabular}
}
\centering\resizebox{0.49\textwidth}{!}{
\begin{tabular}{llllll|l}
\toprule
Food101 & Oxford Pets & DTD & Flowers102 & FGVCAircraft & Stanford Cars & \textbf{Avg}\\
\hline
82.59 & 85.06 & 43.18 & 62.89 & 24.99 & 58.54 & 55.01 \\
\hline
\textbf{82.79} & 86.54 & \textbf{43.99} & \textbf{64.01} & \textbf{26.94} & \textbf{57.08} & \blue{\textbf{56.05}} \\
82.37   & \textbf{88.03}   & 43.35   & 63.62   & 25.77   & 56.21   & 54.74  \\
\bottomrule
\end{tabular}
}
\caption{\textbf{Importance of semantics in DCLIP.} The favorable performance of similarity score averaging (\textit{mean}) over simply selecting the maximum similarity score (\textit{max}), which in some cases can even underperform the CLIP baseline built upon, points to the limited impact of LLM-generated semantics on improved visual classification.}
\label{tab:max_dclip_main}
\end{table}
\begin{table*}[t]
\centering\resizebox{\linewidth}{!}{
\begin{tabular}{l|llllllll|l}
\toprule
\textbf{ViT-B/32} & ImageNetV2 & ImageNet & CUB200 & EuroSAT & Places365 & Food101 & Oxford Pets & DTD & \textbf{Avg}\\
\hline
DCLIP~\cite{dclip} &  \textbf{55.82} & \textbf{63.12} & 52.47 & \textbf{43.29} & 40.47 & 82.79 & 86.54 & \textbf{43.99} & \blue{\textbf{58.56}}\\
\hline
DCLIP (interchanged) & 52.51 \small$\pm0.42$ & 59.62 \small$\pm0.13$ & 52.52 \small$\pm0.41$ & 33.63 \small$\pm4.16$ & 35.52 \small$\pm0.32$ & 81.71 \small$\pm0.35$ & 86.28 \small$\pm0.50$ & 38.42 \small$\pm1.14$ & 55.03 \small$\pm1.56$\\
DCLIP (scrambled) & 55.12 \small$\pm0.12$ & 62.57 \small$\pm0.12$ & 52.18 \small$\pm0.28$ & 40.48 \small$\pm2.52$ & 39.91 \small$\pm0.08$ & 82.46 \small$\pm0.13$ & 86.10 \small$\pm0.40$ & 41.58 \small$\pm0.31$ & 57.55 \small$\pm0.92$\\
DCLIP (random,  1x) & 54.11 \small$\pm0.28$ & 61.37 \small$\pm0.18$ & 52.42 \small$\pm0.19$ & 36.83 \small$\pm4.27$ & 38.80 \small$\pm0.26$ & 82.86 \small$\pm0.23$ & 85.99 \small$\pm0.62$ & 42.20 \small$\pm0.85$ & 56.82 \small$\pm1.57$\\
DCLIP (random,  5x) & 55.43 \small$\pm0.12$ & 62.81 \small$\pm0.05$ & 52.66 \small$\pm0.17$ & 38.57 \small$\pm1.52$ & 40.54 \small$\pm0.05$ & 84.03 \small$\pm0.11$ & 86.75 \small$\pm0.21$ & 43.41 \small$\pm0.74$ & 58.02 \small$\pm0.61$\\
\bottomrule
\end{tabular}
}
\caption{\textbf{Progression from systematic to fully randomized descriptor scrambling.} To model systematic semantic shifts, we randomly swap descriptor lists between classes (\textit{interchanged}), before progressing to shuffling descriptor words within the classes (\textit{scrambled}) and randomly sampling LLM-generated descriptors for each class (\textit{random}) from the complete set of descriptors with counts as in (or five times that of) DCLIP (\textit{1x}, \textit{5x}). 
As can be seen, a systematic shift results in a notable performance drop, while more independently randomized descriptors can recover the DCLIP performance, aligning with the observation that fully randomized prompt averaging is the main performance driver for \texttt{Waffle}CLIP.}
\label{tab:semantic_ablation}
\end{table*}
\begin{figure}[t]
    \centering
    \includegraphics[width=\linewidth]{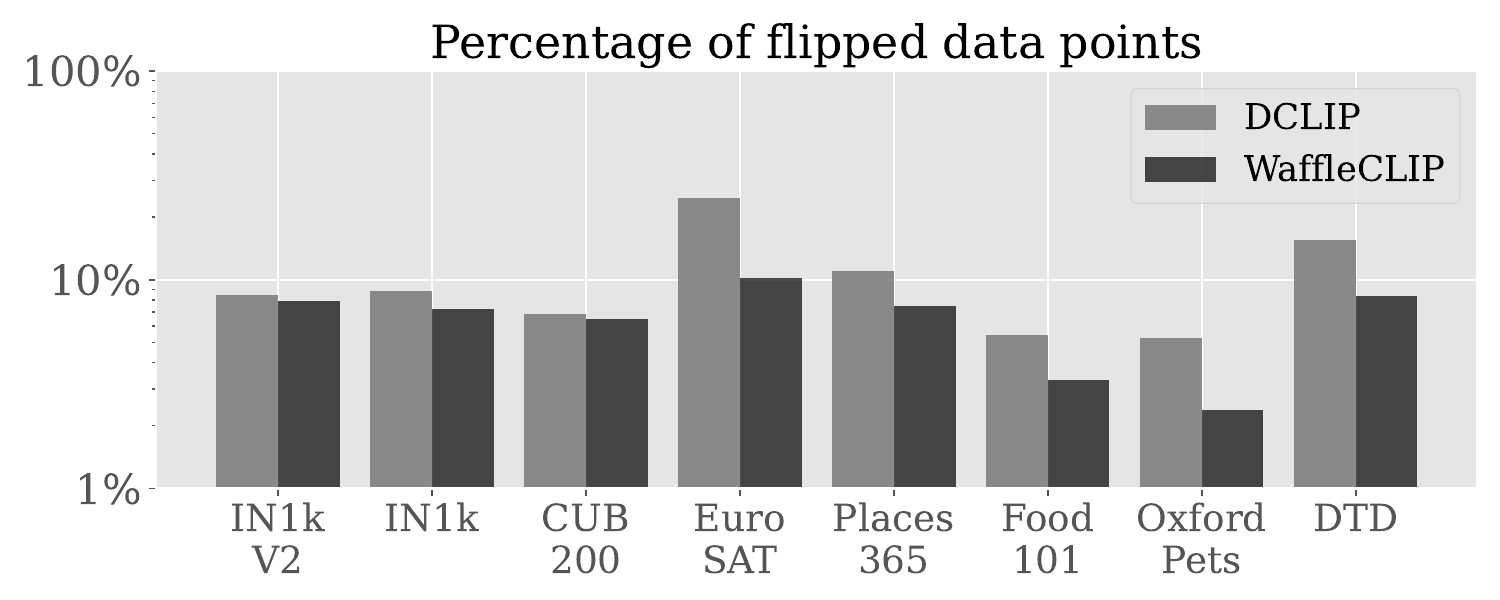} 
    \vspace{-17pt}
    \caption{
    Label flipping experiment from CLIP to DCLIP or \texttt{Waffle}CLIP. 
    Each bar indicates the percentage of data points getting either positively or negatively flipped (i.e. labelled correctly or incorrectly adjusted) when switching from CLIP to either DCLIP or \ourmodel. The consistently higher flip percentage
    indicates structural differences between natural language descriptors and randomized ones.}
    \label{fig:flips}
\end{figure}

We further support this by studying additional descriptor randomization variants beyond those in \S\ref{subsec:method_1}. In particular, instead of swapping specific descriptors, we interchange full class-specific descriptor lists (\textit{interchanged}). As descriptions often contain class-specific keywords, this models a systematic semantic shift away from the actual class. Additionally, we evaluate shuffling words within a descriptor list (\textit{shuffled}), and descriptor lists subsampled from all available ones (\textit{random}). This gives a progression from systematic to more independent descriptor randomization.

Our results in Tab.~\ref{tab:semantic_ablation} reveal that directly interchanging full \textit{class-dependent} descriptor lists (\textit{interchanged}) drops performance significantly (e.g.\ from $58.56\%$ to $55.03\%$ for ViT-B/32). In cases where no such shift is happening, we find performances to match DCLIP (e.g.\ $86.54\%\rightarrow86.28\%$ on Oxford Pets). Similarly, when moving from a systematic shift closer to fully randomized descriptors, performance approaches DCLIP (\textit{scrambled} with $58.56\%\rightarrow57.55\%$ to \textit{random} with $58.56\%\rightarrow 58.02\%$, see supp. material for more results). 

While this offers further evidence for \ourmodel and the fact that class-dependent ensembling drives gains, it does not yet allow us to directly compare the impact on the prediction behavior of LLM-generated descriptors and randomized ones.

\subsubsection{Structural differences between LLM-generated and randomized descriptors} 
We consider the percentages of samples that get positively or negatively flipped - i.e.\ classified correctly while previously being classified incorrectly (and vice versa) -  when moving from CLIP to either DCLIP or \ourmodel in Fig.~\ref{fig:flips}. We find that using LLM-generated fine-grained descriptors flips significantly more predictions than randomized words and characters, even when \texttt{Waffle}CLIP outperforms DCLIP. For example, DCLIP achieves $43.29\%$ compared to \ourmodel with $44.31\%$ on EuroSAT or $82.79\%$ to $83.25\%$ on Food101 in Tab.~\ref{tab:main}, but DCLIP flips a significantly larger portion of samples than \ourmodel.

This reveals that full sentence, LLM-generated descriptors have a \textit{structurally different} impact on the classification process, which we find to be \textit{complementary} to randomization (see Tab.~\ref{tab:main}, \textit{+ GPT descr.}), where the use of both leads to additional improvements over \ourmodel (e.g. $58.57\%\rightarrow 60.21\%$ for ViT-B/32).
%

This means that even if additional semantics are not the leading factor, LLMs for structured descriptor generation can still facilitate more robust class embeddings. Even \textit{with} access to an external model for producing class descriptors, \ourmodel can provide additional benefits.

\begin{figure}[t]
    \centering
    \includegraphics[width=1\linewidth]{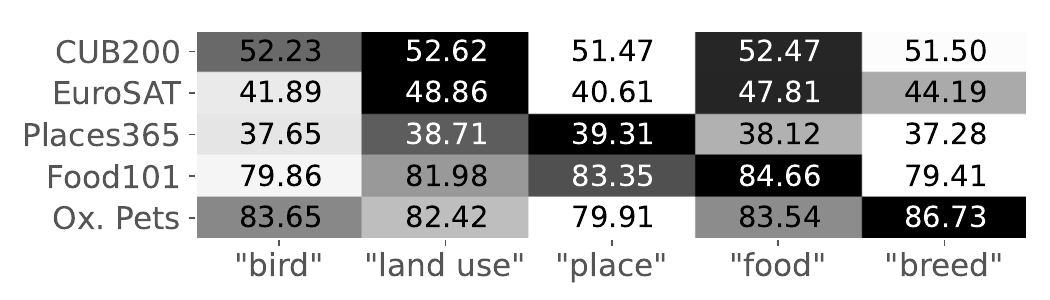} 
    \vspace{-17pt}    
    \caption{Study of the semantic impact of GPT-3 generated high-level concepts through a semantic confusion matrix, where we cycle high-level concepts between each benchmark. We find that interchanging the concepts generally reduces performance, indicating that high-level concepts provide complementary semantic context.
    } 
    \label{fig:concept_transfer}
\end{figure}

\begin{table}[t]
\centering\resizebox{\linewidth}{!}{
\begin{tabular}{l|lll|l}
\toprule
\textbf{ViT-B/32} & Flowers102 & FGVCAircraft & Stanford Cars & \textbf{Avg} \\
\hline
CLIP~\cite{clip} & 62.89 & 24.99 & 58.54 & 48.81\\
DCLIP \cite{dclip} & 64.01 & 26.94 & 57.08 & 49.34\\

\hline
\texttt{Waffle}CLIP & 66.27 \small$\pm0.26$ & 25.66 \small$\pm0.19$ & 58.91 \small$\pm0.17$ & 50.28 \small$\pm0.21$\\
+ Concepts & \textbf{67.19} \small$\pm0.19$ & 28.44 \small$\pm0.22$ & \textbf{59.70} \small$\pm0.12$ & \blue{\textbf{51.78}} \small$\pm0.18$\\
+ GPT dsc. + Conc. & 66.71 \small$\pm0.39$ & \textbf{28.96} \small$\pm0.37$ & 59.33 \small$\pm0.14$ & 51.67 \small$\pm0.32$\\
\bottomrule
\end{tabular}
}
\caption{We find similar performance improvements with \ourmodel and high-level concept guidance for three additional standard visual classification benchmarks not studied in \cite{dclip}, which in parts do not benefit from LLM-generated descriptors (e.g.\ \textit{Stanford Cars}). 
}
\label{tab:additional_benchmarks}
\end{table}
\begin{table}[t]
\centering\resizebox{\linewidth}{!}{
    \centering
    \begin{tabular}{l|c|c|c} 
    \toprule
         \textbf{Benchmarks} & ImageNet-R~\cite{hendrycks2021imagenetr} & ImageNet-S~\cite{wang2019imagenets} & ImageNet-A~\cite{hendrycks2021imageneta} \\
         \midrule
         CLIP~\cite{clip} & 65.97 & 40.73 & 29.63 \\
         DCLIP~\cite{dclip} & 65.12 & 41.09 & 29.19 \\
         \midrule
         \texttt{Waffle}CLIP & \textbf{67.31} & \textbf{42.00} & \textbf{31.52} \\
    \bottomrule
    \end{tabular}}
    \caption{Performance gains of \ourmodel on distribution-shifted datasets to study impacts on OOD generalization. Our results further highlight the general applicability of simple averaging over randomized descriptors, even in cases where the use of natural language ones may fail.}
    \label{tab:dist_shift_comp}
\end{table}

\subsection{Semantic guidance with high-level concepts}\label{subsec:highlevel}
While we verified the relevance of additional semantic context through fine-grained descriptors, methods using additional fine-grained class information~\cite{dclip,chils,cupl} suffer from the inherent ambiguities of some class names.
As proposed in \S\ref{subsec:method_2}, our aim is to understand if high-level semantic context can be used to resolve such ambiguities by providing coarse semantic guidance for the class-retrieval process. 

Our results with extracted high-level concepts
in Tab.~\ref{tab:main}, i.e. (\textit{+ Concepts}), demonstrate consistent and significant improvements across most benchmarks and backbones when used with CLIP, with \texttt{Waffle}CLIP, and even alongside \texttt{Waffle}CLIP and DCLIP. 
These improvements are especially evident on benchmarks with ambiguous (e.g.\ Food101) or generic labeling (e.g.\ EuroSAT, with labels such as \textit{Industrial} or \textit{Residential}): For ViT-B/32, classification accuracy increases from $40.78\%$ to $48.86\%$ when applied to CLIP.

Overall, the average classification accuracy also increases consistently (e.g.\ from $57.34\%$ to $58.96\%$ for ViT-B/32). This even beats DCLIP, despite only being applicable on five out of eight benchmarks ($58.96\%$ versus $58.56\%$).
When applied to \ourmodel, improvements across most benchmark and backbone settings are also significant, although we find diminishing returns on the largest backbone, ViT-L/14, with average performance increasing only from $68.95\%$ to $69.12\%$ (see suppl.\ material). This might be due to its capabilities of retaining the most common concepts associated with specific classes, resulting in a robust class retrieval setup when averaging over multiple randomized descriptor variants. 

We verify the benefits of high-level semantics further by looking at performance changes when concepts are interchanged (Fig.~\ref{fig:concept_transfer}). For most benchmarks, the largest improvements are obtained with GPT-generated concepts. 
Some off-diagonal terms with higher scores, e.g.\ CUB200 where \texttt{"bird"} performs similar to/worse than \texttt{"land use"}/\texttt{"food"}, do appear out of distribution and warrant future research to improve our understanding of how semantics concepts are truly encoded in large VLMs.

However, seeing maximum performances primarily on the diagonal heuristically supports that additional semantics introduced as high-level concepts and commonalities, can offer reliable guidance. Indeed, considering a selection of ambiguous samples such as \texttt{"Boxer"} or \texttt{"Sphynx"} in the Oxford Pets dataset, \texttt{"Mussels"}, \texttt{"Oysters"} or \texttt{"Grilled Salmon"} in the Food101 dataset, or highly generic labels such as \texttt{"Industrial"} or \texttt{"Residential"} in the EuroSAT satellite image dataset, we find a consistent increase in average similarity to all associated test images by up to $13\%$. This confirms that concept guidance can re-align and refine class embeddings based on the relevant context.

\begin{figure}[t]
    \centering
    \includegraphics[width=\linewidth]{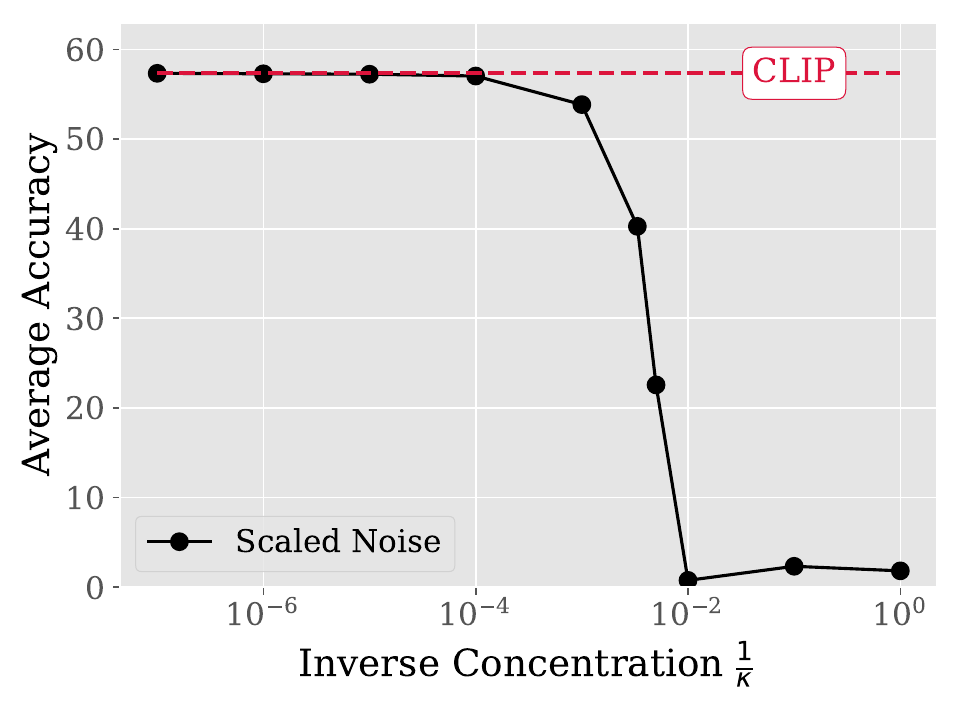} 
    \vspace{-15pt}
    \caption{We evaluate changes in the performance of the CLIP baseline when using latent space noise (modeled using vMF distributions centered around the default, unaltered CLIP language embeddings) and thus also provide an implicit comparison against \ourmodel). However, as can be seen, reducing latent noise (i.e.\ increasing concentration~$\kappa$) converges to initial performance while significantly dropping the performance when increasing the latent noise scale. This highlights no notable benefit of deploying noise in the latent space.}
    \label{fig:noise_scale}
\end{figure}

\begin{table*}[t]
\centering\resizebox{\linewidth}{!}{
\begin{tabular}{l|p{1.5cm}p{1.5cm}p{1.5cm}p{1.5cm}p{1.5cm}p{1.5cm}p{1.5cm}p{1.5cm}p{1.5cm}p{1.5cm}p{1.5cm}|l}
\toprule
\textbf{ViT-B/32} & IN1k-V2 & IN1k & CUB & Euro & Places & Food & Pets & DTD & Flowers & FGVC & Cars & \textbf{Avg}\\
\hline
CLIP~\cite{clip} & 54.71 & 62.01 & 51.28 & 40.78 & 39.12 & 82.59 & 85.06 & 43.18 & 62.89 & 24.99 & 58.54 & 55.01\\
DCLIP~\cite{dclip} & 55.82 & 63.12 & 52.47 & 43.29 & 40.47 & 82.79 & 86.54 & \textbf{43.99} & 64.01 & 26.94 & 57.08 & 56.05 \\
\hline
P. Ensemble & 55.49 \tiny$\pm0.21$ & 62.79 \tiny$\pm0.29$ & 51.46 \tiny$\pm0.43$ & 45.76 \tiny$\pm0.49$ & 40.58 \tiny$\pm0.06$ & 82.67 \tiny$\pm0.37$ & 83.26 \tiny$\pm0.72$ & 42.53 \tiny$\pm0.54$ & 63.30 \tiny$\pm0.33$ & 25.14 \tiny$\pm0.45$ & 58.38 \tiny$\pm0.29$ & 55.58 \tiny$\pm0.42$\\
+ Concepts & $\downarrow$ & $\downarrow$ & \textbf{52.08} \tiny$\pm0.17$ & \textbf{49.80} \tiny$\pm0.66$ & \textbf{40.61} \tiny$\pm0.14$ & \textbf{84.45} \tiny$\pm0.15$ & \textbf{87.42} \tiny$\pm0.20$ & $\downarrow$ & \textbf{65.38} \tiny$\pm0.27$ & \textbf{26.64} \tiny$\pm0.50$ & \textbf{59.12} \tiny$\pm0.14$ & \textbf{56.94} \tiny$\pm0.34$\\
\hline
\ourmodel & \textbf{55.92} \tiny$\pm0.08$ & \textbf{63.31} \tiny$\pm0.09$ & 52.38 \tiny$\pm0.12$ & 44.31 \tiny$\pm1.07$ & 40.56 \tiny$\pm0.07$ & 83.25 \tiny$\pm0.21$ & 85.70 \tiny$\pm0.25$ & 43.16 \tiny$\pm0.25$ & 66.27 \tiny$\pm0.26$ & 25.66 \tiny$\pm0.19$ & 58.91 \tiny$\pm0.17$ & 56.31 \tiny$\pm0.37$\\
+ Concepts & $\downarrow$ & $\downarrow$ & \textbf{52.83} \tiny$\pm0.19$ & \textbf{48.51} \tiny$\pm0.70$ & \textbf{40.97} \tiny$\pm0.08$ & \textbf{85.21} \tiny$\pm0.06$ & \textbf{87.52} \tiny$\pm0.10$ & $\downarrow$ & \textbf{67.19} \tiny$\pm0.19$ & \textbf{28.44} \tiny$\pm0.22$ & \textbf{59.70} \tiny$\pm0.12$ & \blue{\textbf{57.52}} \tiny$\pm0.26$\\
\bottomrule
\end{tabular}
}
\caption{\textbf{Prompt ensembling versus \ourmodel (+concepts)}. Across all visual classification benchmarks, we conduct comparisons to equivalent prompt ensembling, which leverages handcrafted prompts. Our results show matching or improved performance of \ourmodel (improving on eight out of eleven benchmarks), with the increase in average classification performance of \ourmodel compared to prompt ensembling higher than the increase of a prompt-ensembled version over the respective standard CLIP baseline, \textbf{without} requiring a handcrafted list of prompts.}
\label{tab:prompt_ensembling_main}
\end{table*}


\subsection{Evaluation on additional (OOD) benchmarks.}\label{subsec:add_benchmarks}
We observe further evidence for the generality of $\texttt{Waffle}$CLIP and concept guidance by studying three additional benchmarks beyond those in Tab.~\ref{tab:main} and \cite{dclip}:  Flowers102~\cite{flowers102} (extracted concept: \texttt{"flower"}), FGVCAircraft~\cite{fgcvaircraft} (\texttt{"aircraft"}), and StanfordCars~\cite{cars196}  (\texttt{"car"}). 
Our results in Tab.~\ref{tab:additional_benchmarks} (and in the suppl.\ material for other backbones) again show consistent gains when going from CLIP to \texttt{Waffle}CLIP or \ourmodel\textit{+ Concepts}.

Interestingly, DCLIP is detrimental on very fine-grained benchmarks like Stanford Cars, losing $1.46\%$ against CLIP. 
We speculate that this is due to semantically similar descriptors for multiple classes that are coarser than the actual class label (e.g.\ \texttt{"BMW Active Hybrid"} and \texttt{"BMW 1 Series"} being assigned similar generic BMW descriptors). Consequently, embeddings of related classes are systematically moved too close, harming performance.
Meanwhile, \texttt{Waffle}CLIP (\textit{+ Concepts}) can still offer performance boosts ($58.54\%\rightarrow58.91\%\rightarrow59.70\%$).

Furthermore, we study \ourmodel on benchmarks measuring OOD generalization: Adversarial natural images (ImageNet-A, \cite{hendrycks2021imageneta}), sketches (ImageNet-S, \cite{wang2019imagenets}) and renditions (ImageNet-R, \cite{hendrycks2021imagenetr}). Results in Tab.~\ref{tab:dist_shift_comp} show that while DCLIP does not improve consistently, \ourmodel operates well even for out-of-distribution data (e.g. $29.63\%\rightarrow 31.52\%$ on ImageNet-A).

\subsection{Comparison to prompt ensembles}\label{subsec:promptensembles}
We also compare \ourmodel to prompt ensembling (c.f.\ e.g.\ \cite{cupl}) with the same budget of 30 randomly selected prompt options from a list of eighty handcrafted ones (taken from \cite{cupl}, such as \texttt{"A tattoo of a \{class\}."}, \texttt{"A \{class\} in a video game."}, ...). Unlike \ourmodel, prompt ensembling still requires human input and design. Results on all eleven benchmarks are listed in Tab.~\ref{tab:prompt_ensembling_main}, which favor \ourmodel, outperforming prompt ensembling in eight out of eleven benchmarks and comparable performance on the remaining ones.

In particular, improvements over prompt ensembling are higher than the improvement of prompt ensembling over vanilla CLIP ($56.31\%\rightarrow55.58\%\rightarrow55.01\%$).
This further supports the benefit of extracting more robust semantic representations, for which randomized descriptors provide a cheap and suitable tool.

In addition to that, we highlight the complementarity of high-level concept guidance in combination with prompt ensembling in Tab.~\ref{tab:prompt_ensembling_main} (wherever the classname is included, we simply use \texttt{"a \{concept\}: a \{classname\}"} instead), raising the average classification accuracy from $55.58\%$ to $56.94\%$.

\subsection{Comparison to latent noise}\label{subsec:latent_noise}
To highlight that input-level class-conditioned randomization is crucial, we compare our results using \ourmodel to latent random noise applied on the hypersperical representation of CLIP. To model the corresponding noise distribution, we use unimodal von-Mises-Fisher distributions  (as commonly done, c.f. e.g. \cite{wang2020uniform,hyperspherical_vae,zimmermann2021invert,ooniso}) of class embedding vectors $\hat{\phi}^c$:
\begin{equation}
    p(\hat{\phi}^c|\phi^c_l, \kappa) = \mathcal{C}_d(\kappa)\exp(\kappa {\phi^c_l}^T\hat{\phi}^c),
\end{equation}
centered around a class centroid vectors $\phi^c_l$ with constant normalization $\mathcal{C}_d(\kappa)$ only dependent on the dimensionality of $\phi$ and concentration $\kappa$. Note that $\phi^c_l$ is simply the language embedding produced by CLIP when utilizing the unaltered classname and input prompt.
To sample from a vMF distribution around each class embedding, we leverage the sampler utilized in \cite{hyperspherical_vae,Kirchhof2022ANP} with the same budget of 30 noise embeddings. 

Average performance as a function of the (inverse) concentration $\kappa$ is visualized in Fig.~\ref{fig:noise_scale}. 
For high concentrations (i.e.\ random embedding samples placed close to the mean direction), one can replicate the CLIP performance. For higher variances, performance continuously drops, with a hard inflection at around $\kappa\approx10^4$. This shows that class-conditioned randomized descriptors as used in \ourmodel are crucial for providing a more robust estimate of semantic concepts, and cannot be simulated through simple embedding space noise.

\subsection{Ablations}\label{subsec:ablations}
\subsubsection{Dependence on descriptor counts}
\begin{figure}[t]
    \centering
    \includegraphics[width=\linewidth]{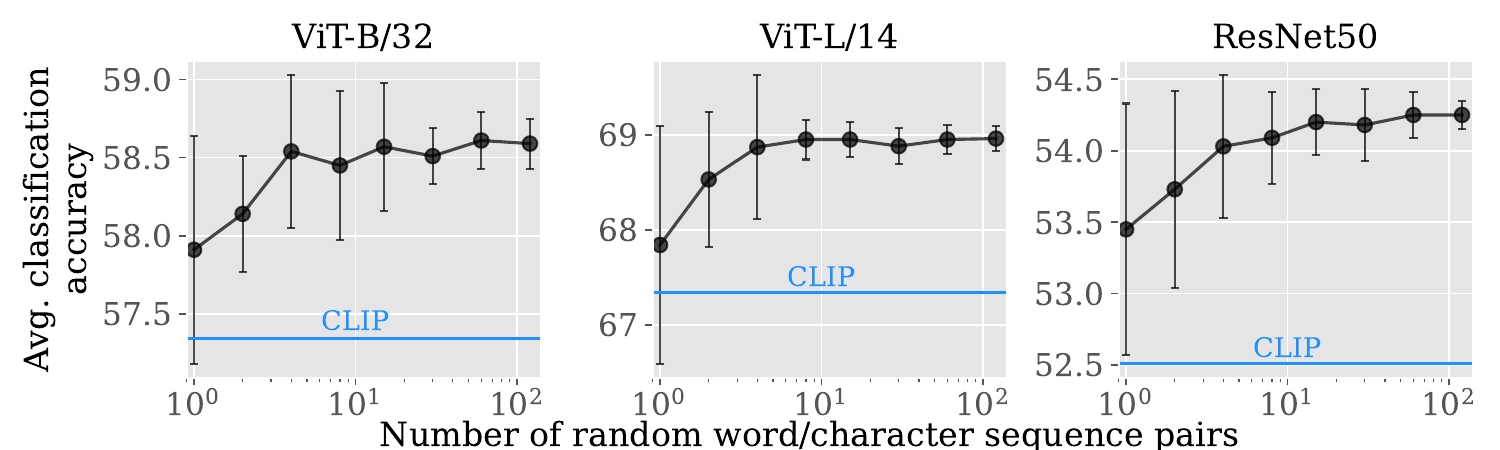} 
    \vspace{-11pt}
    \caption{Ablation study on the number of randomized word and character descriptors used in \texttt{Waffle}CLIP. We find consistent competitive performance gains with just four randomized descriptor pairs (both random words and random characters, c.f.\ DCLIP Tab.\ \ref{tab:main}). Note that the CLIP reference (blue line) is already outperformed with just a single descriptor pair.}  
    \label{fig:random_ablation}
\end{figure}

We study the impact of the randomized word and character sequence pair count for \ourmodel in Fig.~\ref{fig:random_ablation}. A value of one indicates a single pair comprising a random words and a separate random characters descriptor, respectively. As can be seen, we already achieve competitive performance with 4 to 15 descriptor pairs (c.f.\ DCLIP in Tab.~\ref{tab:main}), while consistently outperforming CLIP (blue line) even with a single randomized descriptor pair. As class embeddings can be computed \textit{a priori}, and the generation of random words and characters does not require external model queries, the impact on overall setup and inference time is low, making \texttt{Waffle}CLIP and its extensions very attractive for enhancing image classification performance of VLMs.

\subsubsection{Impact of randomization types}
Finally, we analyze how performance changes when either using only random character sequences or only random word sequences instead of a combination of both as in \ourmodel.
Across benchmarks and architectures (see Tab.~\ref{tab:miniablate}), we observe dichotomies in performance between either random word or random character sequences, often performing either best or worst on a specific benchmark and backbone, while the joint usage of random words and character sequences strikes a consistent and best transferable average improvement across benchmarks and backbone architectures. Therefore, we chose the joint usage of both random words and characters as our default setup.

\begin{table}[t]
\centering\resizebox{\linewidth}{!}{
\begin{tabular}{l|lll}
\toprule
\textbf{Avg.} & ViT-B/32 & ViT-L/14 & RN50\\
\hline
Joint & 58.57 \small$\pm0.41$ & 68.95 \small$\pm0.18$ & 54.20 \small$\pm0.23$\\
Random Words & 58.18 \small$\pm0.44$ & 68.73 \small$\pm0.58$ & 55.24 \small$\pm0.41$\\
Random Characters & 58.59 \small$\pm0.27$ & 68.02 \small$\pm0.14$ & 53.79 \small$\pm0.16$\\
\bottomrule
\end{tabular}
}
\caption{\textbf{Randomized descriptor modes.} We provide a quick comparison between the joint usage of randomized word and character sequences as opposed to their singular usage, i.e. either only randomized word or character sequences. Our experiments show that joint usage provides the most consistent performance improvements across benchmarks and backbones.}
\label{tab:miniablate}
\end{table}

\section{Conclusion} In this work, we systematically examined the benefits of using LLM-generated class descriptors for improved training-free image classification with vision-language models (VLMs). 
In-depth studies reveal how similar performance gains can be achieved by replacing LLM-generated descriptors with randomized ones, giving rise to \ourmodel. Without access to external LLMs, across eleven visual classification benchmarks, we get comparable or better results than those obtained when using fine-grained GPT-3 generated descriptors. This makes \ourmodel very attractive for practical use in true zero-shot scenarios, and it serves as a crucial sanity check for future methods using external queries.
We also show that VLMs struggle to leverage the actual semantics introduced through fine-grained semantic descriptors, and instead show that if given access to external LLMs, semantics are better exploited through coarse, high-level concepts. Using specific queries, we show how these can be automatically extracted, while jointly helping to address issues of class ambiguity.

\section*{Acknowledgements}
This work was supported by DFG project number 276693517, by BMBF FKZ: 01IS18039A, by the ERC (853489 - DEXIM), and by EXC number 2064/1 – project number 390727645.
KR and JMK thank the European Laboratory for Learning and Intelligent Systems (ELLIS) PhD program and the International Max
Planck Research School for Intelligent Systems (IMPRS-IS) for support.

{\small
\bibliographystyle{ieee_fullname}
\bibliography{main}
}

\newpage

\newcommand{\ask}[1]{\textcolor{teal}{ASK: #1}}
\appendix
\begin{table*}[t]
\centering\resizebox{0.98\textwidth}{!}{
\begin{tabular}{l|llllllll|l}
\toprule
\textbf{ViT-L/14} & ImageNetV2 & ImageNet & CUB200 & EuroSAT & Places365 & Food101 & Oxford Pets & DTD & \textbf{Avg}\\
\hline
CLIP~\cite{clip} & 67.90 & 73.37 & 62.24 & 56.03 & 40.46 & 92.55 & 93.30 & 52.87 & 67.34\\
DCLIP~\cite{dclip} & \textbf{69.72} & 75.26 & 63.53 & 58.72 & \textbf{42.60} & 92.81 & 93.89 & \textbf{56.60} & 69.14\\
\hline
DCLIP (same, 1x) & 69.27 \small$\pm0.23$ & 75.05 \small$\pm0.15$ & 64.21 \small$\pm0.36$ & 57.59 \small$\pm1.72$ & 42.01 \small$\pm0.23$ & 93.15 \small$\pm0.13$ & 93.97 \small$\pm0.22$ & 55.16 \small$\pm0.47$ & 68.80 \small$\pm0.66$\\
DCLIP (same, 2x) & 69.58 \small$\pm0.21$ & \textbf{75.30} \small$\pm0.16$ & \textbf{64.30} \small$\pm0.26$ & \textbf{59.32} \small$\pm1.63$ & 42.28 \small$\pm0.17$ & \textbf{93.31} \small$\pm0.05$ & \textbf{94.04} \small$\pm0.11$ & 55.31 \small$\pm0.50$ & \blue{\textbf{69.18}} \small$\pm0.62$\\
\bottomrule
\end{tabular}
}
\centering\resizebox{0.98\textwidth}{!}{
\begin{tabular}{l|llllllll|l}
\toprule
\textbf{ResNet50} & ImageNetV2 & ImageNet & CUB200 & EuroSAT & Places365 & Food101 & Oxford Pets & DTD & \textbf{Avg}\\
\hline
CLIP~\cite{clip} &  51.34 & 58.16 & 45.20 & 28.09 & 36.63 & 78.37 & 83.76 & 38.51 & 52.51\\
DCLIP~\cite{dclip} &  52.70 & 59.66 & \textbf{47.76} & 34.27 & 38.39 & 78.59 & \textbf{85.77} & \textbf{41.01} & 54.77\\
\hline
DCLIP (same, 1x) & 52.63 \small$\pm0.28$ & 59.69 \small$\pm0.30$ & \textbf{47.76} \small$\pm0.39$ & 32.74 \small$\pm1.49$ & 38.63 \small$\pm0.22$ & 80.08 \small$\pm0.58$ & 85.36 \small$\pm0.52$ & 40.77 \small$\pm0.63$ & 54.71 \small$\pm0.67$\\
DCLIP (same, 1x) & \textbf{52.89} \small$\pm0.23$ & \textbf{59.90} \small$\pm0.26$ & 47.70 \small$\pm0.29$ & \textbf{34.37} \small$\pm1.27$ & \textbf{38.93} \small$\pm0.21$ & \textbf{80.11} \small$\pm0.30$ & 85.34 \small$\pm0.29$ & 40.91 \small$\pm0.79$ & \blue{\textbf{55.02}} \small$\pm0.58$\\
\bottomrule
\end{tabular}
}
\caption{\textbf{Motivating random class descriptors - additional backbones.} Extension of our motivational experiments from Tab.~\ref{tab:motivation} with ViT-L/14 and ResNet50 backbones.}
\label{tab:supp_vitl14_motivation}
\end{table*}

\begin{table*}[t]
\centering\resizebox{0.98\textwidth}{!}{
\begin{tabular}{l|llllllll|l}
\toprule
\textbf{ViT-L/14} & ImageNetV2 & ImageNet & CUB200 & EuroSAT & Places365 & Food101 & Oxford Pets & DTD & \textbf{Avg}\\
\hline
CLIP~\cite{clip} & 67.90 & 73.37 & 62.24 & 56.03 & 40.46 & 92.55 & 93.30 & 52.87 & 67.34\\
+ Concepts & $\downarrow$ & $\downarrow$ & 63.01 & 61.23 & 41.07 & 93.52 & 93.65 & $\downarrow$ & 68.32\\
\hline
DCLIP~\cite{dclip} & \textbf{69.72} & 75.26 & 63.53 & 58.72 & 42.60 & 92.81 & 93.89 & \textbf{56.60} & 69.14\\
\hline
\ourmodel (ours) & 69.48 \small$\pm0.08$ & 75.30 \small$\pm0.04$ & \textbf{64.18} \small$\pm0.13$ & \textbf{61.17} \small$\pm0.35$ & 42.26 \small$\pm0.10$ & 93.31 \small$\pm0.09$ & 91.98 \small$\pm0.11$ & 53.94 \small$\pm0.29$ & 68.95 \small$\pm0.18$\\
+ Concepts & $\downarrow$ & $\downarrow$ & 63.40 \small$\pm0.17$ & 60.20 \small$\pm0.87$ & 42.57 \small$\pm0.09$ & \textbf{93.65} \small$\pm0.05$ & \textbf{94.38} \small$\pm0.08$ & $\downarrow$ & 69.12 \small$\pm0.33$\\
+ GPT descr. & \textbf{69.80} \small$\pm0.13$ & \textbf{75.57} \small$\pm0.06$ & \textbf{64.32} \small$\pm0.21$ & \textbf{60.63} \small$\pm1.23$ & \textbf{42.96} \small$\pm0.12$ & 93.28 \small$\pm0.08$ & 93.35 \small$\pm0.22$ & \textbf{56.33} \small$\pm0.42$ & \blue{\textbf{69.53}} \small$\pm0.48$\\
+ GPT descr. + Concepts & $\downarrow$ & $\downarrow$ & 63.14 \small$\pm0.16$ & \textbf{61.82} \small$\pm1.07$ & \textbf{42.95} \small$\pm0.09$ & 93.49 \small$\pm0.04$ & 94.12 \small$\pm0.09$ & $\downarrow$ & \blue{\textbf{69.65}} \small$\pm0.42$\\
\bottomrule
\end{tabular}
}
\centering\resizebox{0.98\textwidth}{!}{
\begin{tabular}{l|llllllll|l}
\toprule
\textbf{ResNet50} & ImageNetV2 & ImageNet & CUB200 & EuroSAT & Places365 & Food101 & Oxford Pets & DTD & \textbf{Avg}\\
\hline
CLIP~\cite{clip} & 51.34 & 58.16 & 45.20 & 28.09 & 36.63 & 78.37 & 83.76 & 38.51 & 52.51\\
+ Concepts & $\downarrow$ & $\downarrow$ & 46.60 & 34.06 & 37.43 & 80.89 & 83.43 & $\downarrow$ & 53.80\\
\hline
DCLIP~\cite{dclip} &  \textbf{52.70} & 59.66 & 47.76 & 34.27 & 38.39 & 78.59 & \textbf{85.77} & \textbf{41.01} & 54.77\\
\ourmodel (ours) & \textbf{52.89} \small$\pm0.15$ & \textbf{60.12} \small$\pm0.12$ & 47.68 \small$\pm0.15$ & 31.34 \small$\pm0.47$ & 38.32 \small$\pm0.10$ & 79.68 \small$\pm0.17$ & 84.32 \small$\pm0.20$ & 39.25 \small$\pm0.27$ & 54.20 \small$\pm0.23$\\
+ Concepts & $\downarrow$ & $\downarrow$ & \textbf{48.34} \small$\pm0.13$ & 35.08 \small$\pm0.42$ & 39.03 \small$\pm0.08$ & \textbf{81.38} \small$\pm0.08$ & \textbf{85.80} \small$\pm0.12$ & $\downarrow$ & 55.24 \small$\pm0.21$\\
+ GPT descr. + Concepts & $\downarrow$ & $\downarrow$ & \textbf{48.41} \small$\pm0.21$ & \textbf{37.36} \small$\pm0.62$ & \textbf{39.43} \small$\pm0.07$ & 81.17 \small$\pm0.09$ & \textbf{85.82} \small$\pm0.16$ & $\downarrow$ & \blue{\textbf{55.75}} \small$\pm0.26$\\
\bottomrule
\end{tabular}
}
\caption{\textbf{Performance of \ourmodel with additional backbones.} Here, we extend the comparison of \ourmodel (Tab.~\ref{tab:main}) to GPT-generated fine-grained class descriptors in DCLIP \cite{dclip} for ViT-L/14 and ResNet50 backbones. We find similarly consistent insights, where our LLM-free \ourmodel can match the performance of DCLIP. Joint usage of both randomized and LLM-generated descriptors again reveals complementarity (\ourmodel \textit{+ GPT descr}). In addition to that, the usage of automatically extracted high-level semantic concepts can provide consistent additional performance gains (\textit{+ Concepts}). We use $(\downarrow)$ to denote the same results as previous lines where high-level concept guidance is not applicable.}
\label{tab:suppp_main}
\end{table*}

\begin{table*}[t]
\centering\resizebox{0.98\textwidth}{!}{
\begin{tabular}{l|llllllll|l}
\toprule
\textbf{ViT-L/14} & ImageNetV2 & ImageNet & CUB200 & EuroSAT & Places365 & Food101 & Oxford Pets & DTD & \textbf{Avg}\\
\hline
DCLIP~\cite{dclip} & \textbf{69.72} & \textbf{75.26} & 63.53 & \textbf{58.72} & \textbf{42.60} & 92.81 & 93.89 & \textbf{56.60} & \blue{\textbf{69.14}}\\
\hline
DCLIP (interchanged) & 66.44 \small$\pm0.12$ & 72.07 \small$\pm0.15$ & 63.62 \small$\pm0.44$ & 51.49 \small$\pm4.89$ & 37.06 \small$\pm0.41$ & 91.30 \small$\pm0.30$ & 93.74 \small$\pm0.28$ & 49.84 \small$\pm0.78$ & 65.69 \small$\pm1.77$\\
DCLIP (scrambled) & 68.68 \small$\pm0.21$ & 74.47 \small$\pm0.11$ & 63.78 \small$\pm0.13$ & 55.98 \small$\pm2.01$ & 41.29 \small$\pm0.23$ & 92.29 \small$\pm0.20$ & 93.52 \small$\pm0.18$ & 53.28 \small$\pm1.12$ & 67.91 \small$\pm0.83$\\
DCLIP (random,  1x) &68.01 \small$\pm0.22$ & 73.89 \small$\pm0.08$ & 63.81 \small$\pm0.22$ & 55.72 \small$\pm2.01$ & 40.32 \small$\pm0.29$ & 92.37 \small$\pm0.31$ & 93.60 \small$\pm0.19$ & 52.83 \small$\pm0.46$ & 67.57 \small$\pm0.76$\\
DCLIP (random,  5x) & 69.27 \small$\pm0.17$ & 75.11 \small$\pm0.08$ & 64.25 \small$\pm0.16$ & 58.34 \small$\pm1.55$ & 42.11 \small$\pm0.14$ & 93.22 \small$\pm0.12$ & 93.88 \small$\pm0.09$ & 55.28 \small$\pm0.23$ & 68.93 \small$\pm0.57$\\
\bottomrule
\end{tabular}
}
\centering\resizebox{0.98\textwidth}{!}{
\begin{tabular}{l|llllllll|l}
\toprule
\textbf{ResNet50} & ImageNetV2 & ImageNet & CUB200 & EuroSAT & Places365 & Food101 & Oxford Pets & DTD & \textbf{Avg}\\
\hline
DCLIP~\cite{dclip} &  52.70 & 59.66 & \textbf{47.76} & 34.27 & 38.39 & 78.59 & \textbf{85.77} & \textbf{41.01} & 54.77\\
\hline
DCLIP (interchanged) & 49.80 \small$\pm0.22$ & 56.35 \small$\pm0.06$ & 47.68 \small$\pm0.32$ & 28.17 \small$\pm4.43$ & 33.77 \small$\pm0.34$ & 77.59 \small$\pm0.29$ & 84.60 \small$\pm0.63$ & 35.81 \small$\pm1.12$ & 51.72 \small$\pm1.64$\\
DCLIP (scrambled) & 52.20 \small$\pm0.20$ & 59.21 \small$\pm0.06$ & 47.60 \small$\pm0.39$ & 34.98 \small$\pm2.00$ & 37.90 \small$\pm0.18$ & 78.33 \small$\pm0.14$ & 85.07 \small$\pm0.34$ & 39.19 \small$\pm0.95$ & 54.31 \small$\pm0.81$\\
DCLIP (random,  1x) &51.60 \small$\pm0.29$ & 58.29 \small$\pm0.15$ & 47.37 \small$\pm0.23$ & 30.18 \small$\pm4.18$ & 36.82 \small$\pm0.26$ & 78.87 \small$\pm0.24$ & 84.52 \small$\pm0.17$ & 38.89 \small$\pm0.85$ & 53.32 \small$\pm1.52$\\
DCLIP (random,  5x) & 52.81 \small$\pm0.09$ & 59.73 \small$\pm0.05$ & 47.74 \small$\pm0.10$ & 34.53 \small$\pm0.74$ & 38.62 \small$\pm0.15$ & 80.20 \small$\pm0.13$ & 85.30 \small$\pm0.15$ & 40.29 \small$\pm0.46$ & 54.90 \small$\pm0.32$\\
\bottomrule
\end{tabular}
}
\caption{\textbf{Progression from systematic to fully randomized descriptor scrambling - additional backbones.} We extend our descriptor scrambling progression studies from Tab.~\ref{tab:semantic_ablation} to two additional backbones: ViT-L/14 and ResNet50. In both cases, the same trend can be seen, in which a move from systematic semantic shift to independently subsampled descriptors can recover the performance of DCLIP after an initial performance drop.}
\label{tab:semantic_ablation_supp}
\end{table*}


\section*{\Large{\textsc{Supplementary material:}\\Waffling around for Performance:\\Visual Classification with Random\\Words and Broad Concepts}}
\begin{figure*}[t]
    \centering
    \includegraphics[width=\linewidth]{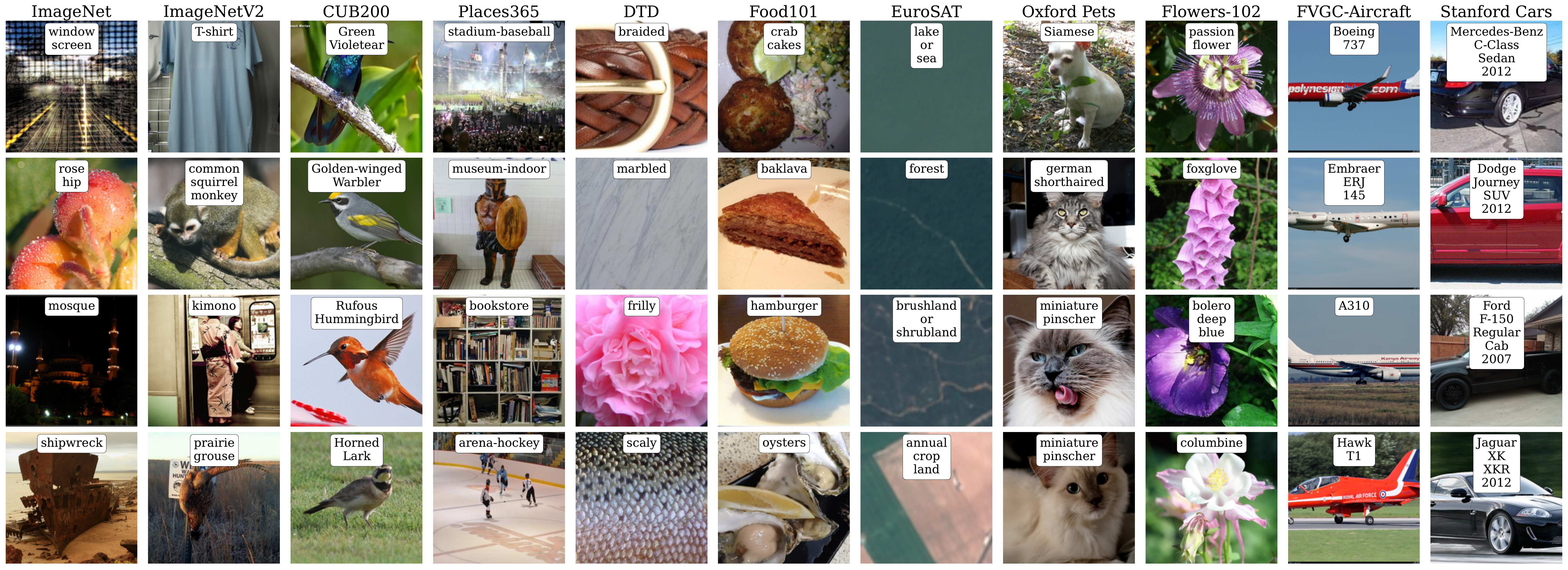} 
    \vspace{-10pt}
    \caption{To get an intuition of the different visual classification tasks, we showcase samples of four randomly selected classes for each of the eleven utilized visual classification benchmarks.}
    \label{fig:sample_images}
\end{figure*}

In this supplementary material, we first provide a collection of additional results in \S\ref{sup:additional_benchmarks} which extend those presented in the main paper to more backbone models. Finally, we showcase the GPT-generated descriptors for our additionally used benchmarks beyond \cite{dclip} (\S\ref{sup:add_descriptors}), and present some exemplary images from the eleven benchmarks used in this work in Fig.~\ref{fig:sample_images}.\\

\section{Additional results}\label{sup:additional_benchmarks}

\mypara{Motivational experiments for random class descriptors.} In Tab.~\ref{tab:supp_vitl14_motivation}, we extend our motivational experiments on random class descriptor assignment to motivate \ourmodel from Tab.~\ref{tab:motivation}, highlighting similar behaviour on both a larger ViT-L/14 and a ResNet50 backbone network. Descriptor randomization does not result in a significant drop in performance, but rather yields performances that match DCLIP.\\

\mypara{Comparison of \ourmodel and DCLIP.} Tab.~\ref{tab:suppp_main} extends results from Tab.~\ref{tab:main} on the ViT-L/14 and ResNet50 backbones, in which \ourmodel as a standalone method, as well as equipped with high-level concepts and/or joint usage of LLM-generated descriptors, is compared to DCLIP. The results confirm our conclusions drawn in \S\ref{subsec:random_descriptors}, wherein \ourmodel, without access to any external LLM, can match the performance of LLM-descriptor-based approaches like DCLIP. In addition to that, we again find complementarity of randomized descriptors and LLM-generated descriptors. Furthermore, we observe performance gains through the usage of automatically generated high-level concepts.\\

\mypara{Progression from systematic to fully randomized descriptor scrambling.} Tab.~\ref{tab:semantic_ablation_supp} extends the descriptor scrambling progression studies from Tab.~\ref{tab:semantic_ablation} to two additional backbones, namely, ViT-L/14 and ResNet50. Similar to the ViT-B/32 backbone, a move from systematic semantic shifts to independently subsampled descriptors can recover and even beat the performance of DCLIP.

\section{Exemplary GPT-3 generated descriptors for additional benchmarks}\label{sup:add_descriptors}
As we introduce descriptions for three additional datasets beyond those used in \cite{dclip}, we provide four example descriptors for three random classes in each dataset.
\subsection*{Flowers102}
\textbf{Pink Primrose}
\begin{itemize}
    \item \texttt{"delicate flower"}
    \item \texttt{"five petals in a star shape"}
    \item \texttt{"pink in color"}
    \item \texttt{"often has yellow center"}
\end{itemize}

\textbf{Balloon Flower}
\begin{itemize}
    \item \texttt{"a delicate flower with five petals"}
    \item \texttt{"a unique balloon-like shape"}
    \item \texttt{"a star-shaped center in the middle of the flower"}
    \item \texttt{"vibrant colors such as pink, purple, blue, white, and yellow"}
\end{itemize}

\textbf{Sunflower}
\begin{itemize}
    \item \texttt{"large, bright yellow petals"}
    \item \texttt{"a dark center surrounded by disk florets"} 
    \item \texttt{"long stem"}
    \item \texttt{"a single, long, narrow leaves tapered to a point"}
\end{itemize}

\subsection*{FGVCAircraft}

\textbf{A300}
\begin{itemize}
    \item \texttt{"black or silver color"}
    \item \texttt{"a rectangular body with rounded edges"}
    \item \texttt{"two lens ports"}
    \item \texttt{"a mode dial"}
\end{itemize}

\textbf{EMB-120}
\begin{itemize}
    \item \texttt{"a cabin with 30-33 seats"}
    \item \texttt{"a distinctive high-wing design"}
    \item \texttt{"two Pratt and Whitney PW118 turboprop engines"}
    \item \texttt{"a T-tail configuration"}
\end{itemize}

\textbf{Tornado}
\begin{itemize}
    \item \texttt{"dark, rotating funnel-shaped cloud"}
    \item \texttt{"strong winds"}
    \item \texttt{"dark clouds"}
    \item \texttt{"heavy precipitation"}
\end{itemize}

\subsection*{Stanford Cars}

\textbf{Acura TL Sedan 2012}
\begin{itemize}
    \item \texttt{"silver, grey, or black exterior"}
    \item \texttt{"Acura logo on the front grille"}
    \item \texttt{"distinctive headlights"}
    \item \texttt{"chrome accents on the exterior"}
\end{itemize}

\textbf{BMW X6 SUV 2012}
\begin{itemize}
    \item \texttt{"four-door SUV"}
    \item \texttt{"sloping roof-line"}
    \item \texttt{"signature BMW kidney grille"}
    \item \texttt{"round headlights and taillights"}
\end{itemize}

\textbf{Honda Odyssey Minivan 2012}
\begin{itemize}
    \item \texttt{"four doors and a hatchback"}
    \item \texttt{"a curved hood"}
    \item \texttt{"wide, round headlights"}
    \item \texttt{"a Honda logo"}
\end{itemize}

\end{document}